\documentclass[lettersize,journal]{IEEEtran}
\usepackage{amsmath,amsfonts}
\usepackage{algorithmic}
\usepackage{algorithm}
\usepackage{array}
\usepackage[caption=false,font=normalsize,labelfont=sf,textfont=sf]{subfig}
\usepackage{textcomp}
\usepackage{stfloats}
\usepackage{url}
\usepackage{verbatim}
\usepackage{graphicx}
\usepackage{cite}
\usepackage{hyperref}
\begin{document}

\title{SSORN: Self-Supervised Outlier Removal Network for Robust Homography Estimation}

\author{
Yi Li\IEEEauthorrefmark{1},
Wenjie Pei\IEEEauthorrefmark{1},
Zhenyu He\IEEEauthorrefmark{2}\\
School of Computer Science and Technology, Harbin Institute of Technology, Shenzhen, China.\\
ly\_res@163.com, wenjiecoder@outlook.com, zhenyuhe@hit.edu.cn.\\
\thanks{$^{*}$ Equal contribution. $\dag$ Corresponding author.}
}



\maketitle

\begin{abstract}
The traditional homography estimation pipeline consists of four main steps: feature detection, feature matching, outlier removal and transformation estimation. Recent deep learning models intend to address the homography estimation problem using a single convolutional network. While these models are trained in an end-to-end fashion to simplify the homography estimation problem, they lack the feature matching step and/or the outlier removal step, which are important steps in the traditional homography estimation pipeline. In this paper, we attempt to build a deep learning model that mimics all four steps in the traditional homography estimation pipeline. In particular, the feature matching step is implemented using the cost volume technique. To remove outliers in the cost volume, we treat this outlier removal problem as a denoising problem and propose a novel self-supervised loss to solve the problem. Extensive experiments on synthetic and real datasets demonstrate that the proposed model outperforms existing deep learning models.
\end{abstract}


\section{Introduction}
Homography estimation is a fundamental task in many computer vision applications, such as Augmented Reality (AR)~\cite{07ParallelTracking,2008ARSurvey,tang20193d}, image stitching~\cite{2017CLKN,chung2019bi,nie2021depth,xue2021stable}, and Simultaneous Location And Mapping (SLAM)~\cite{15ORB-SLAM,2017ORB-SLAM2,shao2021mofisslam}. Although this task has been extensively studied in the past, designing a robust homography estimation method remains a challenging problem.

The traditional homography estimation pipeline mainly includes four steps: feature detection, feature matching, outlier removal and transformation estimation. Among these four steps, feature detection is generally considered to be the most important step. Therefore, a large number of popular handcrafted features, such as SIFT~\cite{2004SIFT}, SURF~\cite{2006SURF} and ORB~\cite{2011ORB}, have been proposed over the past few decades. However, the feature representation capabilities of these feature are limited, especially compared to deep learning features.

Currently, there are two research lines using deep convolutional networks (CNNs) for homography estimation. The first research line is to replace one of the four steps in the traditional homography estimation pipeline with CNNs or neural networks~\cite{ye2019cdbin,18Superpoint,17DSAC,1981RANSAC,zhao2021probabilistic}. For example, SuperPoint~\cite{18Superpoint} adopts a VGG-like architecture for interest point detection and description. SuperGlue~\cite{2020SuperGlue} solves the matching problem based on a  graph neural network. DSAC~\cite{17DSAC} uses a neural networks to mimic the RANSAC algorithm~\cite{1981RANSAC}. However, optimizing each step individually does not necessarily improve the performance of the homography estimation task.

Another research line is to build an end-to-end model for homography estimation. DeTone \textit{et al.}~\cite{2016HomographyNet} propose HomographyNet, the first end-to-end model for homography estimation. It adopts a typical CNNs with 8 convolutional layers and 2 fully connected layers, where the convolutional layers are served as the feature detector/extractor and the fully connected layers are served as the homography estimator. However, it lacks the feature matching and outlier removal steps, which are essential in the tradition homography estimation pipeline. Li \textit{et al.}~\cite{2020SRHEN} propose the SRHEN model, which explicitly implements the feature matching step through a correspondence layer. SRHEN performs much better than HomographyNet, which proves the importance of the feature matching step. However, the output of the correspondence layer in SRHEN contains many outliers. It is quite arduous for a homography estimator consisting of only fully connected layers to reject these outliers on its own.

In this paper, we propose a Self-Supervised Outlier Removal Network (SSORN) for robust homography estimation. The model is built upon a Siamese structure. It consists of four components, \textit{i.e.} a feature extractor, a feature matching module, an outlier removal module, and a homography estimator, to mimic all four steps in the traditional homography estimation pipeline. The model first projects two input images into the same deep space and obtains two feature maps. Then, the feature matching module, which is based on the cost volume technique~\cite{2018PWC}, computes a cost volume between the extracted feature maps. The cost volume generated by the feature matching module usually contains many outliers. To remove the outliers in the cost volume, we treat this outlier removal problem as a denoising problem and propose a novel self-supervised loss to train the outlier removal module. Finally, the homography estimator learns the mapping from a clean cost volume produced by the outlier removal module to the homography matrix. Table~\ref{tab:diff} summaries the difference between our model and existing deep learning models. Experiments on both synthetic and real dataset show that the proposed model outperforms existing deep learning models.

\begin{table*}[!t]
\caption{Difference between our model and existing deep learning models. Our model is the only one that contains all four steps in the traditional homography estimation pipeline.}
\centering
\begin{tabular}{|c||c|c|c|c|}
\hline
	& Feature Extractor & Feature Matching & Outlier Removal & Homography Estimator \\
\hline
	HomographyNet~\cite{2016HomographyNet} & \checkmark & & & \checkmark \\
\hline
	UnsupervisedNet~\cite{2018UnsupervisedNet} & \checkmark & & & \checkmark \\
\hline
	Zhang et al.~\cite{2020Content} & \checkmark & & & \checkmark \\
\hline
	Koguciuk et al.~\cite{2021Perceptual} & \checkmark & & & \checkmark \\
\hline
	SRHEN~\cite{2020SRHEN} & \checkmark & \checkmark & & \checkmark \\
\hline
	Ours & \checkmark & \checkmark & \checkmark & \checkmark \\
\hline
\end{tabular}
\label{tab:diff}
\end{table*}

\begin{figure*}[!t]
\centering
\includegraphics[width=0.99\linewidth]{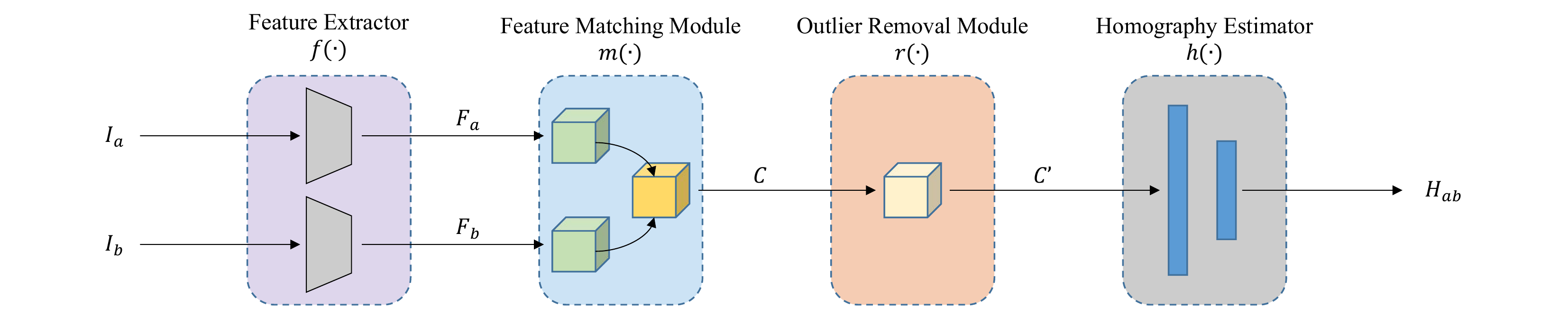}
\caption{The overall architecture of the proposed model. The model is built upon a Siamese structure. It mimics the traditional homography estimation pipeline using four components: a feature extractor, a feature matching module, an outlier removal module and a homography estimator.}
\label{fig:architecture}
\end{figure*}

\section{Related Work}
In this section, we mainly focus on end-to-end homography estimation models. These models can be divided into supervised models and unsupervised models.

DeTone \textit{et al.}~\cite{2016HomographyNet} proposes the first supervised deep learning model, named HomographyNet, for homography estimation. Their model employs a VGG-like network structure with eight outputs on top (corresponding to eight parameters of a homography represented by the 4-point parameterization~\cite{2006Parameterization}). It takes a concatenated image as input and regresses the homography directly. They also propose a data generation rule for generating training and test datasets based on the COCO dataset~\cite{2014COCO} for training and evaluating homography estimation models. Japkowicz \textit{et al.}~\cite{2017HierarchicalNet} proposes a hierarchical model consisting of multiple DNN models with the same structure to handle large transformations. Each DNN model is trained using image pairs with different scale of transformations, enabling the entire model to perform homography estimation in a coarse-to-fine fashion. Li \textit{et al.}~\cite{2020SRHEN} adopts a similar coarse-to-fine framework, but they use different feature maps at different stages and construct multi-level correspondence layers to mimic feature matching step. The above models are designed to solve the general homography estimation problem, while there are other models designed for specific scenarios, such as dynamic scenes~\cite{2020DynamicScenes}, cross-resolution scenes~\cite{2021LocalTrans} and scenes with multiple planes~\cite{nie2021depth}.

Nguyen \textit{et al.}~\cite{2018UnsupervisedNet} treats the homography estimation problem as an unsupervised learning task. The predicted homography, as an intermediate output variable of their model, is used to warp one of the input images. Then, they compute the pixel-wise intensity loss between the warped image and another input image. By minimizing this loss, the model can be trained without ground truth labels. However, the pixel-level intensity loss is very sensitive to illumination changes. For this reason, Zhang \textit{et al.}~\cite{2020Content} proposes to compute the loss in the feature space rather than the intensity space. Koguciuk \textit{et al.}~\cite{2021Perceptual} further extends this idea by computing the perceptual loss~\cite{2016PerceptualLoss}, which significantly improves the robustness of their model against illumination changes. Ye \textit{et al.}~\cite{2021Motion} proposes to use a homography flow, instead of the commonly used 4-point parameterization, as an intermediate representation of the homography. While their model produces good results on image pairs with small viewpoint changes, it cannot handle image pairs with large viewpoint changes.

\section{SSORN}
\subsection{Network Architecture}
Our model is built upon a Siamese structure. It takes two grayscale images $I_{a}$ and $I_{b}$ as input, and produces a homography matrix $H_{ab}$ from $I_{a}$ to $I_{b}$ as output. The entire architecture is composed of four components: a feature extractor $f(\cdot)$, a feature matching module $m(\cdot)$, an outlier removal module $r(\cdot)$, and a homography estimator $h(\cdot)$. The resulting model is illustrated in Fig.~\ref{fig:architecture}.

\noindent\textbf{Feature extractor.} The backbone of the feature extractor follows the ResNet-34 structure. We use the output of \textit{layer2} in the ResNet-34. For a pair of input images $I_{a}$ and $I_{b}$ of size $H \times W \times 1$, the feature extractor $f(\cdot)$ produces feature maps $F_{a}$ and $F_{b}$ of size $H/8 \times W/8 \times C$:
\begin{equation}
    F_{a} = f(I_{a}), \quad F_{b} = f(I_{b})
\end{equation}

\noindent\textbf{Feature matching module.} Since homography is characterized by pixel-to-pixel correspondences rather than image features~\cite{2000MULTIPLE}, feature matching (or establishing correspondences) is an important step in homography estimation. However, the way of establishing pixel-to-pixel correspondences in the traditional homography estimation pipeline is a non-differentiable operation. An alternative solution for feature matching in deep models is the cost volume technique, which has been widely used in optical flow estimation~\cite{2018PWC}. The feature matching module $m(\cdot)$ adopts the cost volume technique to bridge the gap between image features and homography. It computes a cost volume $C$ between the extracted feature maps $F_{a}$ and $F_{b}$:
\begin{equation}
    C = m(F_{a}, F_{b})
\end{equation}

\noindent\textbf{Outlier removal module.} Removing outliers is crucial for robust homography estimation. Traditional outlier removal algorithms like RANSAC~\cite{1981RANSAC} or more recent algorithms like DSAC~\cite{17DSAC} are designed to deal with outliers in pixel-to-pixel correspondences. Unfortunately, these algorithms are not able to handle outliers in a cost volume. In this paper, we treat the problem of removing outliers in a cost volume as a denoising problem and solve it using an outlier removal module. The outlier removal model $r(\cdot)$ takes the cost volume $C$ as input and produces a new cost volume $C'$ as output:
\begin{equation}
    C' = r(C)
\end{equation}

\noindent\textbf{Homography estimator.} The homography estimator simply consists of two fully connected layers. It learns the mapping function from the cost volume $C'$ to the homography matrix $H_{ab}$:
\begin{equation}
    H_{ab} = h(C')
\end{equation}

\begin{figure*}[!t]
\centering
\includegraphics[width=0.99\linewidth]{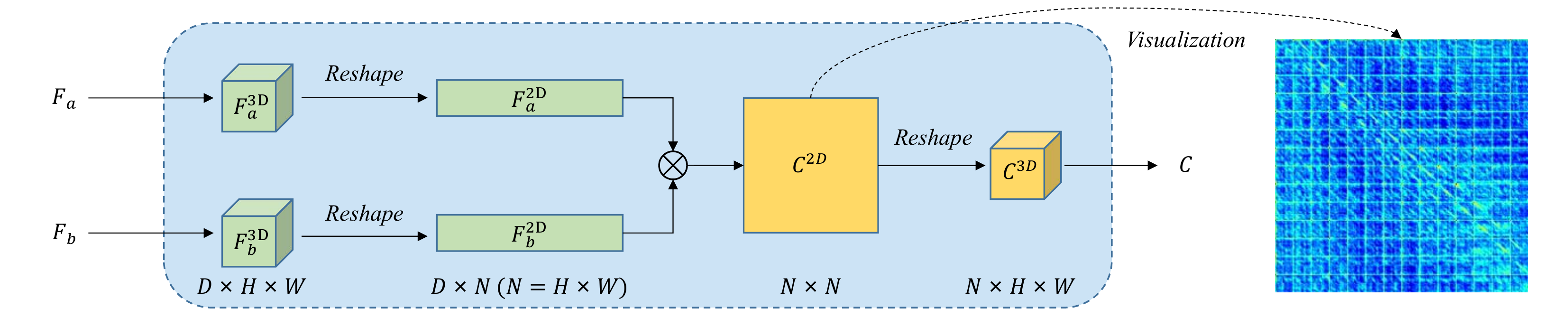}
\caption{The feature matching module adopts the cost volume technique. It explicitly computes a cost volume between two feature maps. We show an example of the cost volume in 2D form on the right.}
\label{fig:costvolume}
\end{figure*}

\begin{figure*}[!t]
\centering
\includegraphics[width=0.99\linewidth]{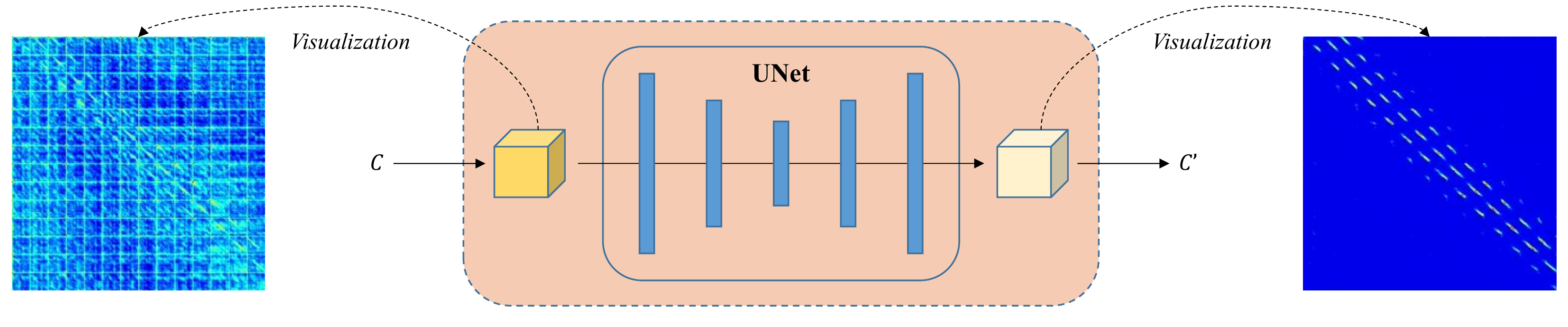}
\caption{The outlier removal model uses the UNet structure as the backbone. It successfully removes most of the outliers in the original cost volume $C$ produced by the feature matching module, obtaining a clean cost volume $C'$.}
\label{fig:denoise}
\end{figure*}

\subsection{Feature Matching Module}
Unlike other components in our model, the feature matching module has no trainable parameters. It explicitly computes a cost volume $C$ between two feature maps $F_{a}$ and $F_{b}$. Intuitively, the cost volume can be thought of as a similarity matrix in 3D form. It stores dense feature matching costs of two sets of feature vectors. The process of computing the cost volume is shown in Fig.~\ref{fig:costvolume}. For clarity, we use the superscripts ``2D'' and ``3D'' to denote 2D and 3D tensors, respectively.

To compute the 3D cost volume $C^{\rm{3D}}$ from $F_{a}^{\rm{3D}}$ to $F_{b}^{\rm{3D}}$, we first reshape $F_{a}^{\rm{3D}}$ and $F_{b}^{\rm{3D}}$ into corresponding 2D tensors $F_{a}^{\rm{2D}}$ and $F_{b}^{\rm{2D}}$ respectively. Then, the matching cost $C^{\rm{2D}}(i,j)$ between the $i-$th feature vector in $F_{a}^{\rm{2D}}$ and the $j-$th feature vector in $F_{b}^{\rm{2D}}$ is implemented as the correlation between the feature vectors:
\begin{equation}
	C^{\rm{2D}}(i,j) = \frac{1}{D}(F_{a}^{\rm{2D}}(i))^{T} \odot F_{b}^{\rm{2D}}(j)
\label{eq:costvolume2}
\end{equation}
where $\odot$ denotes the dot product, $T$ is the transpose operator, and $D$ is the dimension of feature vectors. Accordingly, the full cost volume $C^{\rm{2D}}$ is defined as:
\begin{equation}
	C^{\rm{2D}} = \frac{1}{D}(F_{a}^{\rm{2D}})^{T} \otimes F_{b}^{\rm{2D}}
\label{eq:costvolume3}
\end{equation}
where $\otimes$ denotes the matrix product. Finally, the 2D cost volume $C^{\rm{2D}}$ will be reshaped into the corresponding 3D cost volume $C^{\rm{3D}}$.

\subsection{Outlier Removal Module}
In the traditional view, outliers usually mean incorrect correspondences. In a similar spirit, we can treat those incorrect matching costs in a cost volume as outliers in the cost volume. Since a cost volume stores dense matching costs, it often contains a large number of outliers due to the existence of many similar image features. It is difficult for the homography estimator to predict an exact homography matrix from this cost volume. To this end, we propose the outlier removal model, which treats the outlier removal problem as a cost volume denoising problem. We believe that, for a specific homography matrix, there exists a specific pattern (\textit{i.e.} a clean cost volume) corresponding to that homography matrix. The goal of the outlier removal module is to recover this pattern from the noisy cost volume produced by the feature matching module.

Apparently, the cost volume is different from natural images, thus the priors for the natural image denoising problem are not suitable for the cost volume denoising problem. To address the training problem, we propose a novel self-supervised loss function as described in Sec.~\ref{sec:loss}. Theoretically, any deep denoising model can be used as the outlier removal module. In practice, we adopt the UNet structure as the backbone of the outlier removal model. After training with the proposed self-supervised loss function, the outlier removal module is able to remove most of the outliers in the original cost volume $C$ produced by the feature matching module, obtaining a clean cost volume $C'$, as shown in Fig.~\ref{fig:denoise}. As expected, for a specific homography matrix, the outlier removal module successfully finds a specific pattern (\textit{i.e.} the clean cost volume $C'$) corresponding to that homography matrix.

\begin{figure*}[!t]
\centering
\includegraphics[width=0.99\linewidth]{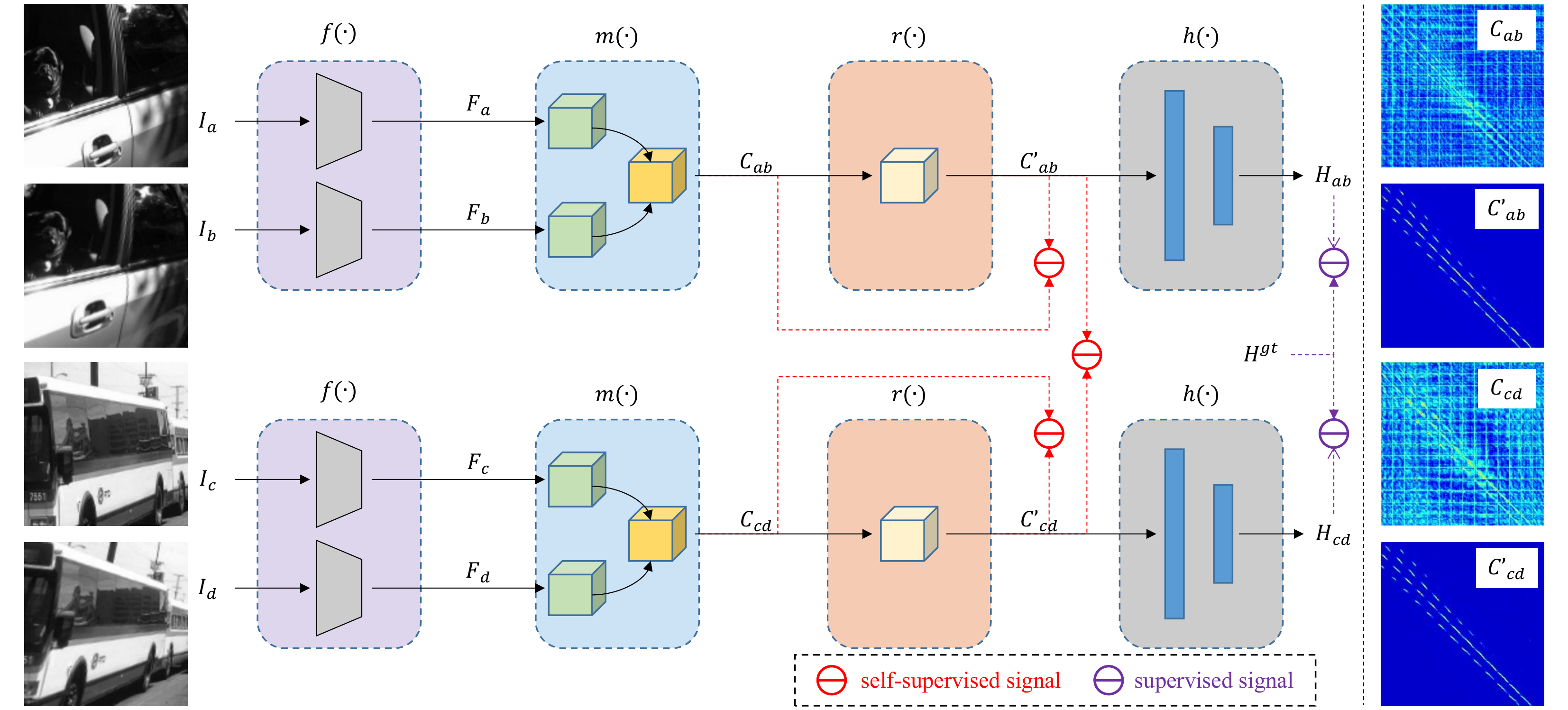}
\caption{The training process of the whole model. Our model requires two image pairs instead of one image pair for training. The two image pairs are related by the same ground truth homography matrix $H^{gt}$. The training loss consists of three self-supervised signals (Eq.~\ref{eq:lss}) and two supervised signals (Eq.~\ref{eq:ls}). For these two image pairs, the two cost volumes $C_{ab}$ and $C_{cd}$ directly produced by the feature matching module may look very different, but the two clean cost volumes $C_{ab}'$ and $C_{cd}'$ produced by the outlier removal module always look the same.}
\label{fig:training}
\end{figure*}

\subsection{Self-Supervised Training}
\label{sec:loss}
Given a pair of images $I_{a}$ and $I_{b}$ and the ground truth homography matrix $H_{ab}^{gt}$, one simple way to train the model is to use the following supervised loss:
\begin{equation}
    L = \| H_{ab} - H_{ab}^{gt} \|_{2}^{2}
\end{equation}
where $\| \cdot \|_{2}^{2}$ is L2 norm.

However, this way does not take full advantage of the outlier removal module, because the above loss function does not ensure that the outlier removal module learns the ability to remove outliers in the cost volume. To this end, we propose an additional self-supervised loss function to train the outlier removal module. We exploit the fact that two image pairs related by the same homography matrix should correspond to a similar clean cost volume. We force the outlier removal module to produce similar outputs for these two image pairs. Fig.~\ref{fig:training} illustrates the training process of the whole model.

Formally, given two image pairs $(I_{a}, I_{b})$ and $(I_{c}, I_{d})$ related by the same ground truth homography matrix $H^{gt}$, the self-supervised loss for training the outlier removal module is designed as:
\begin{equation}
\begin{aligned}
    L_{ss} = \; & \lambda_{1}\|C_{ab}'-C_{cd}'\|_{1} \\
		  & + \lambda_{2}(\|C_{ab}'-C_{ab}\|_{1} + \|C_{cd}'-C_{cd}\|_{1})
\end{aligned}
\label{eq:lss}
\end{equation}
where $\|\cdot\|_{1}$ is L1 norm. The first half of the equation forces the outlier removal module to produce similar cost volumes, and the second half of the equation is used to avoid trivial solutions,~\textit{i.e.} $C_{ab}'=C_{cd}'=0$.

To train the homography estimator, we still need the following supervised loss:
\begin{equation}
    L_{s} = \|H_{ab}-H^{gt}\|_{2}^{2} + \|H_{cd}-H^{gt}\|_{2}^{2}
\label{eq:ls}
\end{equation}

Combining the above two losses, we get the final loss for training the whole model:
\begin{equation}
    L_{f} = L_{s} + L_{ss}
\end{equation}

\subsection{Discussion}
If the idea that a specific homography corresponds to a specific clean cost volume is true, then we seem to be able to derive the grund truth cost volume directly from a ground truth homography, and thus allowing for simple supervised training. In theory, given a ground truth homography and the coordinate of a pixel on one image, we can always compute the coordinate of the unique corresponding pixel on another image, from which we can derive the ground truth cost volume. However, this derivation does not take into account the local self-similarity property of images, that is, adjacent pixels usually have similar (deep) features. In other words, a pixel on one image often has multiple high-scoring matches on another image, not just one match (see the visualization of the clean cost volume in Fig.~\ref{fig:denoise}). It is not trivial to take the local self-similarity property of images into account when deriving the ground truth cost volume from a ground truth homography. This motivates us to propose the self-supervised training, which avoids the difficulty of directly deriving the ground truth cost volume.

\section{Experiments}
\begin{figure*}[t]
\centering
\includegraphics[width=0.98\linewidth]{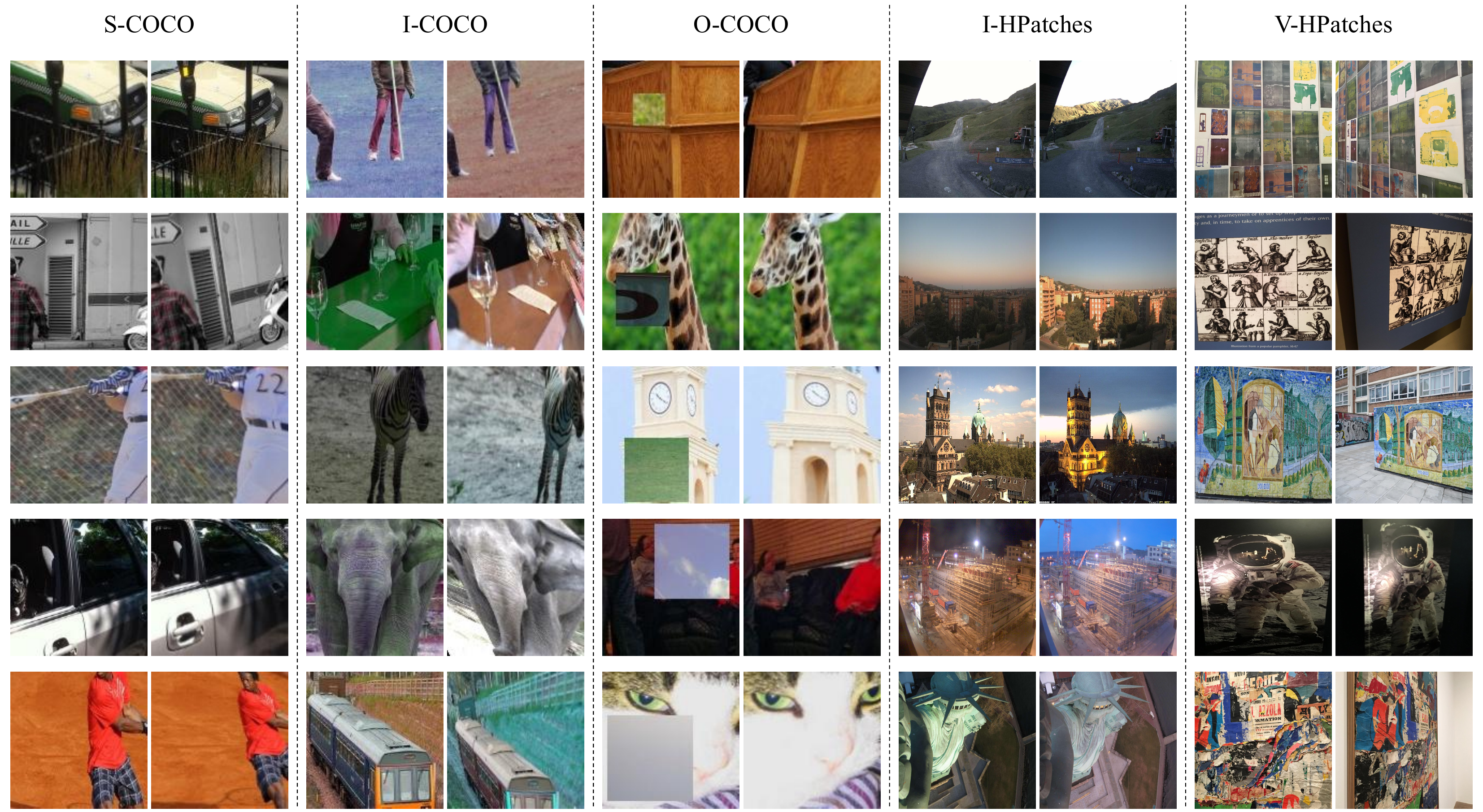}
\caption{Some examples from the datasets used in experiments. Each example consists of two images related by a specific homography matrix. The goal of homography estimation is to predict the homography matrix between the input image pair.}
\label{fig:examples}
\end{figure*}

\subsection{Datasets and Evaluation Metric}
Although there exist several datasets~\cite{2005comparison,2017hpatches,2021image} for homography estimation, these datasets are too small for training deep homography estimation models. To this end, Detone \textit{et al.}~\cite{2016HomographyNet} proposed a synthetic dataset S-COCO based on the COCO dataset~\cite{2014COCO}. S-COCO is generated by applying random projective transformations to the COCO dataset. By using the same data generation rule, we can easily generate two image pairs with the same homography matrix for training our model.

Based on S-COCO, we additionally generate two datasets, called I-COCO and O-COCO, for testing the robustness of different models to illumination and occlusion. In I-COCO, we apply the photometric distortion techniques in~\cite{2016SSD} to S-COCO. The illumination change is controlled by a parameter $\delta$ ranging from 8 to 32 with an interval of 8. In O-COCO, we simulate occlusion by overlaying a random image patch of size $p$ on one of the input images, where $p$ ranges from 0 to 80 with an interval of 10.

In addition to the above three synthetic datasets, we also conduct experiments on a real dataset, HPatches~\cite{2017hpatches}. The dataset contains two subsets: the illumination subset (\textit{i.e.} I-HPatches) and the viewpoint subset (\textit{i.e.} V-HPatches). As this dataset only contains 580 pairs of images, we only use it for testing.

Fig.~\ref{fig:examples} shows some examples from the above datasets. All images are resized to $256 \times 256$ for training and testing. To measure the performance of different models, we adopt the Mean Average Corner Error (MACE)~\cite{2016HomographyNet} as the evaluation metric.

\subsection{Experimental Setup}
We compare our model with 6 recently proposed deep learning models (\textit{i.e.} HomographyNet~\cite{2016HomographyNet}, UnsupervisedNet~\cite{2018UnsupervisedNet}, SRHEN~\cite{2020SRHEN}, Zhang's \textit{et al.}~\cite{2020Content} and Koguciuk's \textit{et al.}~\cite{2021Perceptual} models) and two representative keypoint based methods (\textit{i.e.} SIFT~\cite{2004SIFT} and SuperPoint~\cite{18Superpoint}).

To ensure that the performance difference between deep learning methods comes from better network structure and loss function design, rather than better CNN backbone, we implement these methods to use the same ResNet34 structure as the backbone. Using the same backbone and the same training settings, most deep learning models achieve very similar performance, but this is not the case for SRHEN and our model (see experiments in the following sections). For keypoint based methods, we compare two feature matchers: the nearest neighbor (NN) with Lowe's ratio test (ratio is set to 0.9) and SuperGlue (SG). MAGSAC++~\cite{2019magsac} is used as the robust estimator.

We adopt the 4-point parameterization~\cite{2006Parameterization} to represent the homography matrix instead of $3 \times 3$ matrix form. The weight parameters $\lambda_{1}$ and $\lambda_{2}$ in Eq.~\ref{eq:lss} are empirically set to 0.5 and 0.25, respectively.

\begin{figure*}[t]
\centering
\includegraphics[width=0.98\linewidth]{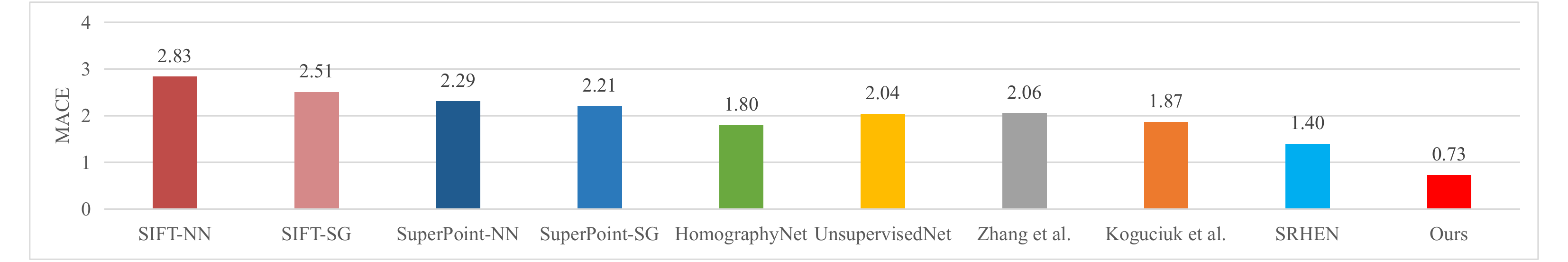}
\caption{MACE for different methods on the S-COCO dataset. Lower MACE means better performance. NN - nearest neighbor. SG - SuperGlue.}
\label{fig:scocoresults0}
\end{figure*}

\begin{figure*}[t]
\centering
\includegraphics[width=0.98\linewidth]{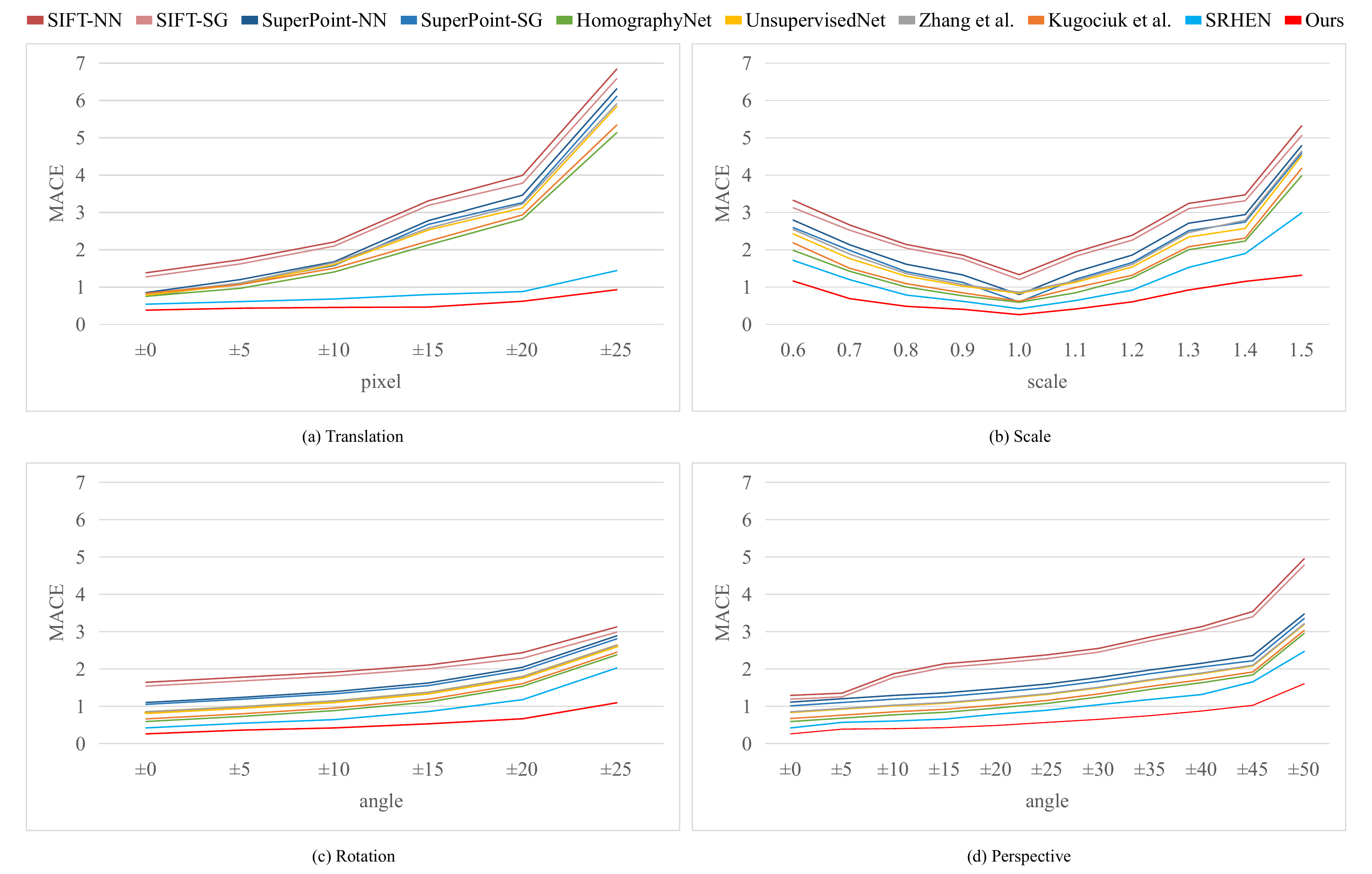}
\caption{Results of different methods under four types of geometric transformations: (a) Translation, (b) Scale, (c) Rotation, and (d) Perspective.}
\label{fig:scocoresults}
\end{figure*}

\subsection{Results on S-COCO Dataset}
We first conduct experiments on the S-COCO dataset. Fig.~\ref{fig:scocoresults0} reports the MACE of different methods on this dataset. In general, deep learning methods outperform keypoint based methods on this dataset, mainly due to the powerful feature representation capability of deep learning features. Compared with those deep models without the feature matching module, SRHEN with the feature matching module achieves a better result. This demonstrates the necessity of the feature matching module not only in the traditional homography estimation pipeline but also in deep homography estimation models. Our model contains not only the feature matching module but also an outlier removal module. It leverages the strength of the traditional homography estimation pipeline and the strength of deep learning features. It obtains a much better result than SRHEN on the S-COCO dataset, which indicates that the outlier removal module is as important as the feature matching module for deep homography estimation models.

To analyze the robustness of different models to different geometric transformations, we generated four subsets according to the data generation rule in~\cite{2016HomographyNet}. Each subset corresponds to a specific type of transformation, namely translation, scale, rotation and perspective transformaitons. We plot the results in Fig.~\ref{fig:scocoresults}. It manifests that all methods perform well when the magnitude of the geometric transformation is small. However, the performance degrades as the magnitude of the geometric transformation increases. Specifically, deep learning models that do not employ the feature matching module perform poorly for large geometric transformations, especially for large translations. Overall, our model is more robust to these four types of geometric transformations than other methods.

\begin{figure*}[!t]
\centering
\includegraphics[width=0.98\linewidth]{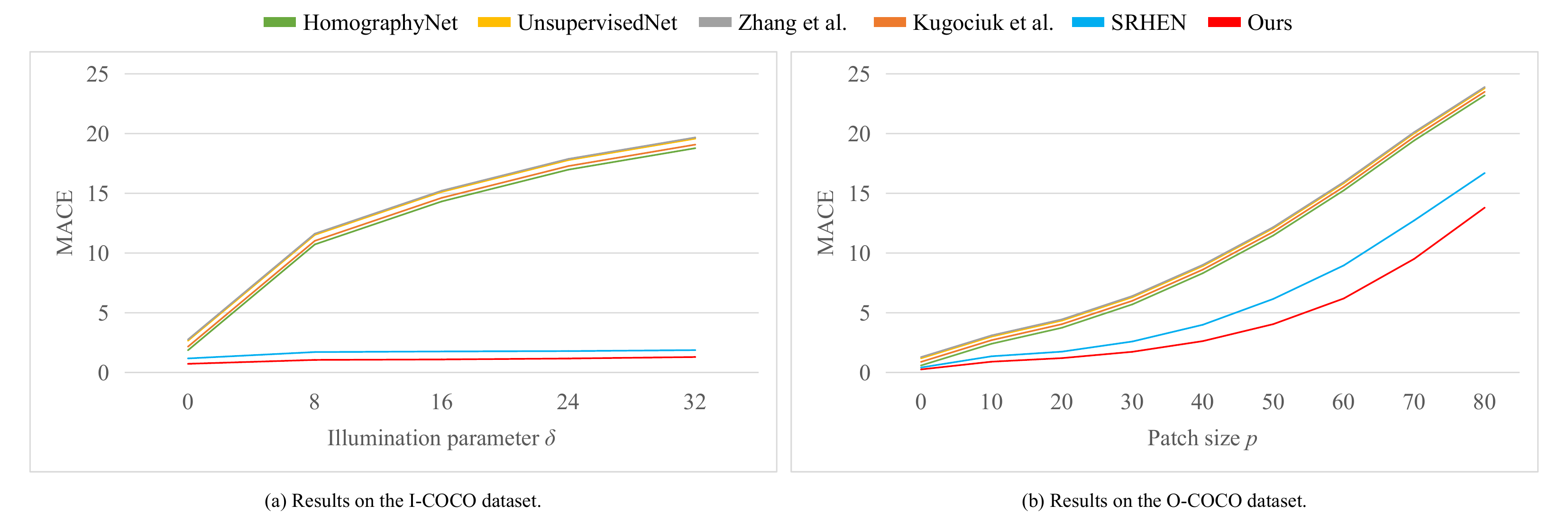}
\caption{Results of the first experimental configuration: all models are trained on the S-COCO dataset and tested on the (a) I-COCO dataset and (b) O-COCO dataset.}
\label{fig:iococoresults1}
\end{figure*}

\begin{figure*}[!t]
\centering
\includegraphics[width=0.99\linewidth]{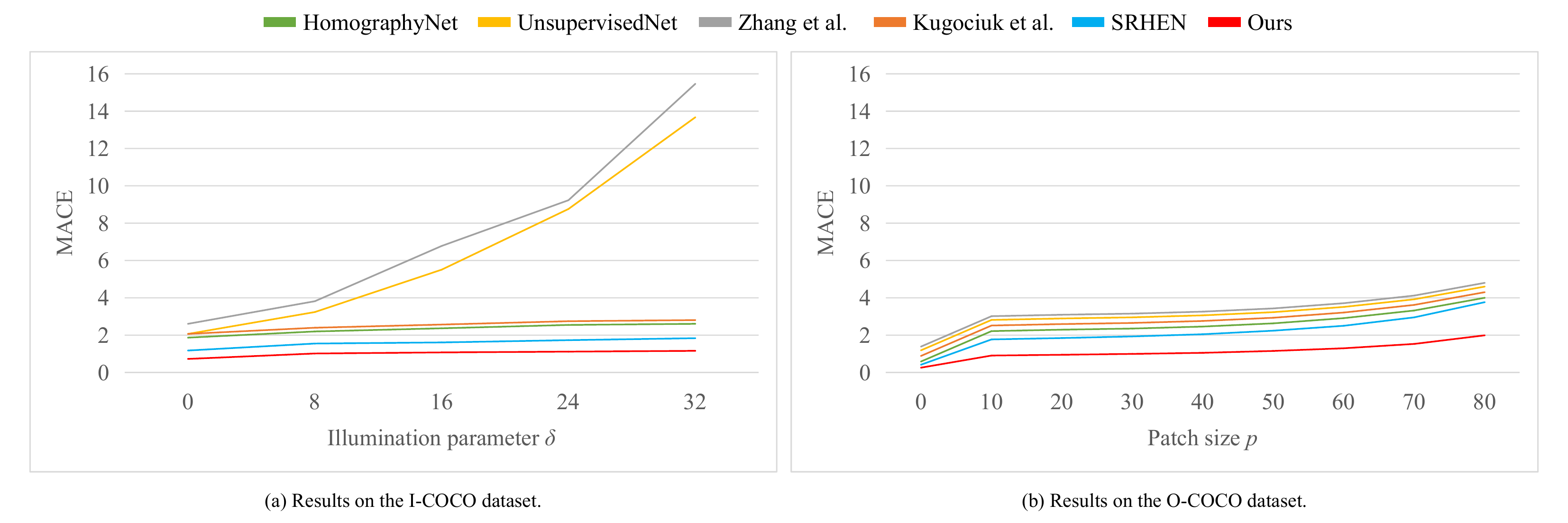}
\caption{Results of the second experimental configuration: all models are (a) trained and tested on the I-COCO dataset and (b) trained and tested on the O-COCO dataset.}
\label{fig:iococoresults2}
\end{figure*}

\subsection{Robustness to Occlusion and Illumination}
To investigate the robustness or the generalization ability of deep learning models to illumination and occlusion, we conduct experiments on the I-COCO and O-COCO datasets. We employ two experimental configurations. In the first experiment, we train all deep learning models on the S-COCO dataset and test them on the I-COCO/O-COCO dataset. In the second experiment, all models are trained and tested on the I-COCO/O-COCO dataset. The experimental results are shown in Fig.~\ref{fig:iococoresults1} and Fig.~\ref{fig:iococoresults2}.

\begin{figure*}[!t]
\centering
\includegraphics[width=0.98\linewidth]{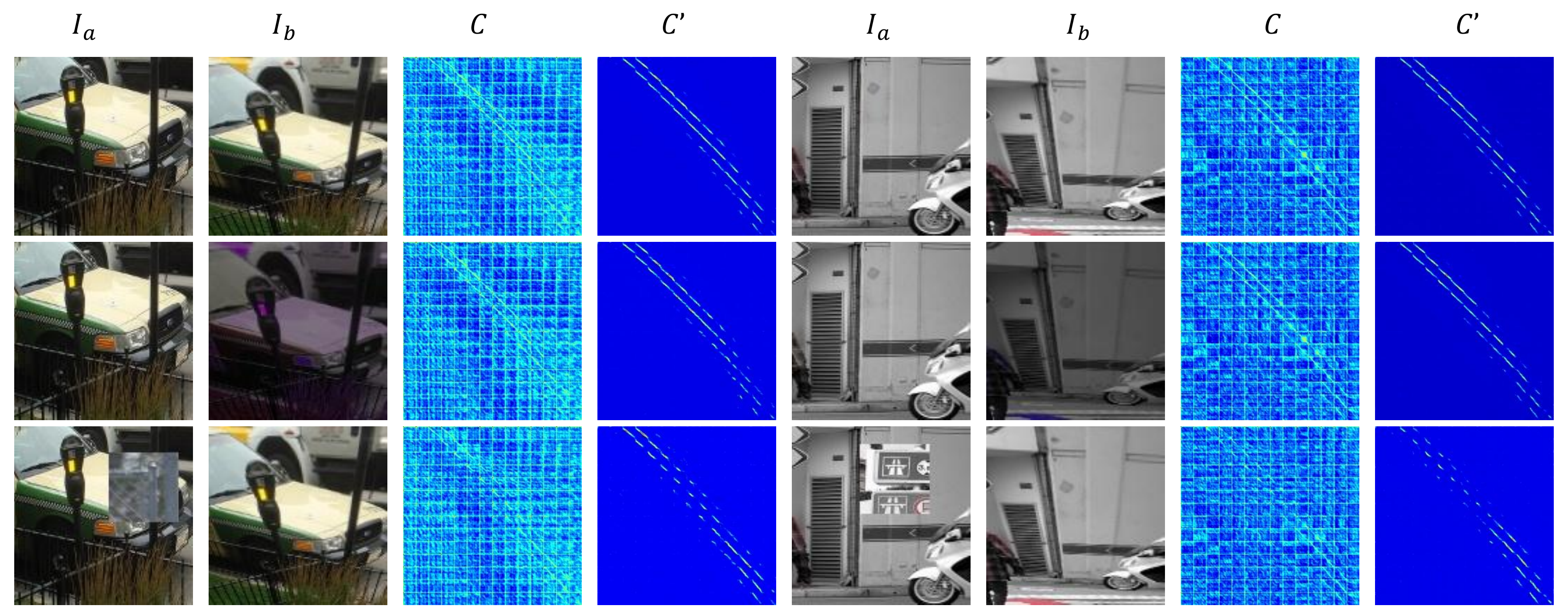}
\caption{All image pairs in this figure are related by the same homography matrix. While the cost volume $C$ produced by the feature matching module may be different due to illumination changes (the second row) or occlusions (the third row), the outlier removal model is able to remove most of the outliers in $C$ and produce the ``correct'' clean cost volume $C'$.}
\label{fig:costvolumeanalysis}
\end{figure*}

Interestingly, we observe that the deep learning models (\textit{i.e.} HomographyNet, UnsupervisedNet, Zhang's \textit{et al.} and Kugociuk's \textit{et al.} models) without the feature matching module obtain significantly different results under the two experimental configurations. This demonstrates that learning the mapping function directly from image features to the homography matrix may lead to the generalization problem. In contrast, SRHEN and our model achieve more stable results under these two experimental configurations. Compared to SRHEN, our model is more robust to illumination and occlusion, as the outlier removal module is able to recover the ``correct'' clean cost volume in most cases (see Fig.~\ref{fig:costvolumeanalysis}).

\subsection{Results on HPatches Dataset}
\begin{figure*}[!t]
\centering
\includegraphics[width=0.98\linewidth]{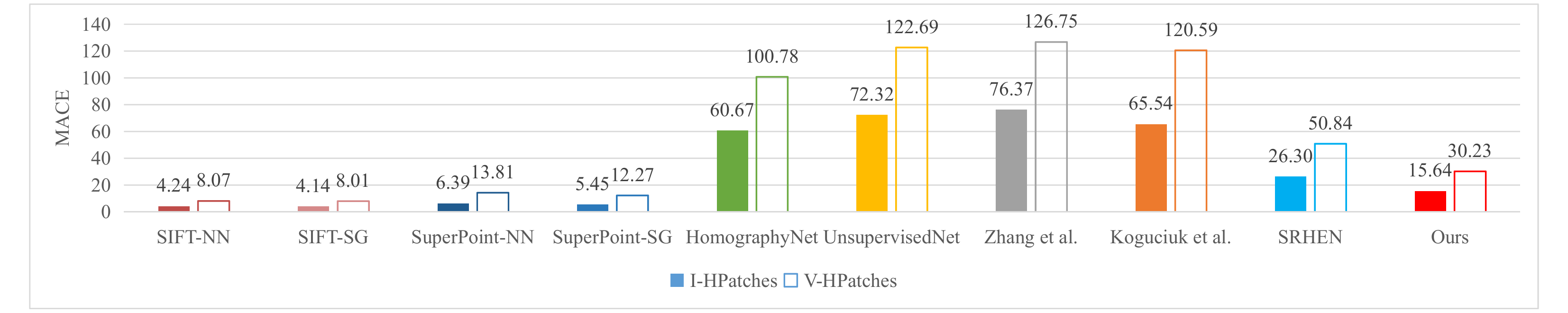}
\caption{MACE for different methods on the HPatches dataset. Lower MACE means better performance.}
\label{fig:hpatchesresults}
\end{figure*}

In the previous sections, we have tested different methods on three synthetic datasets. In order to investigate the generalization ability of these methods to real images, we conduct experiments on the HPatches dataset. Since there are no large-scale real dataset for training the deep learning models, we train all the models on the S-COCO dataset and test them on the HPatches dataset. The experimental results are reported in Fig.~\ref{fig:hpatchesresults}.

\begin{figure*}[!t]
\centering
\includegraphics[width=0.98\linewidth]{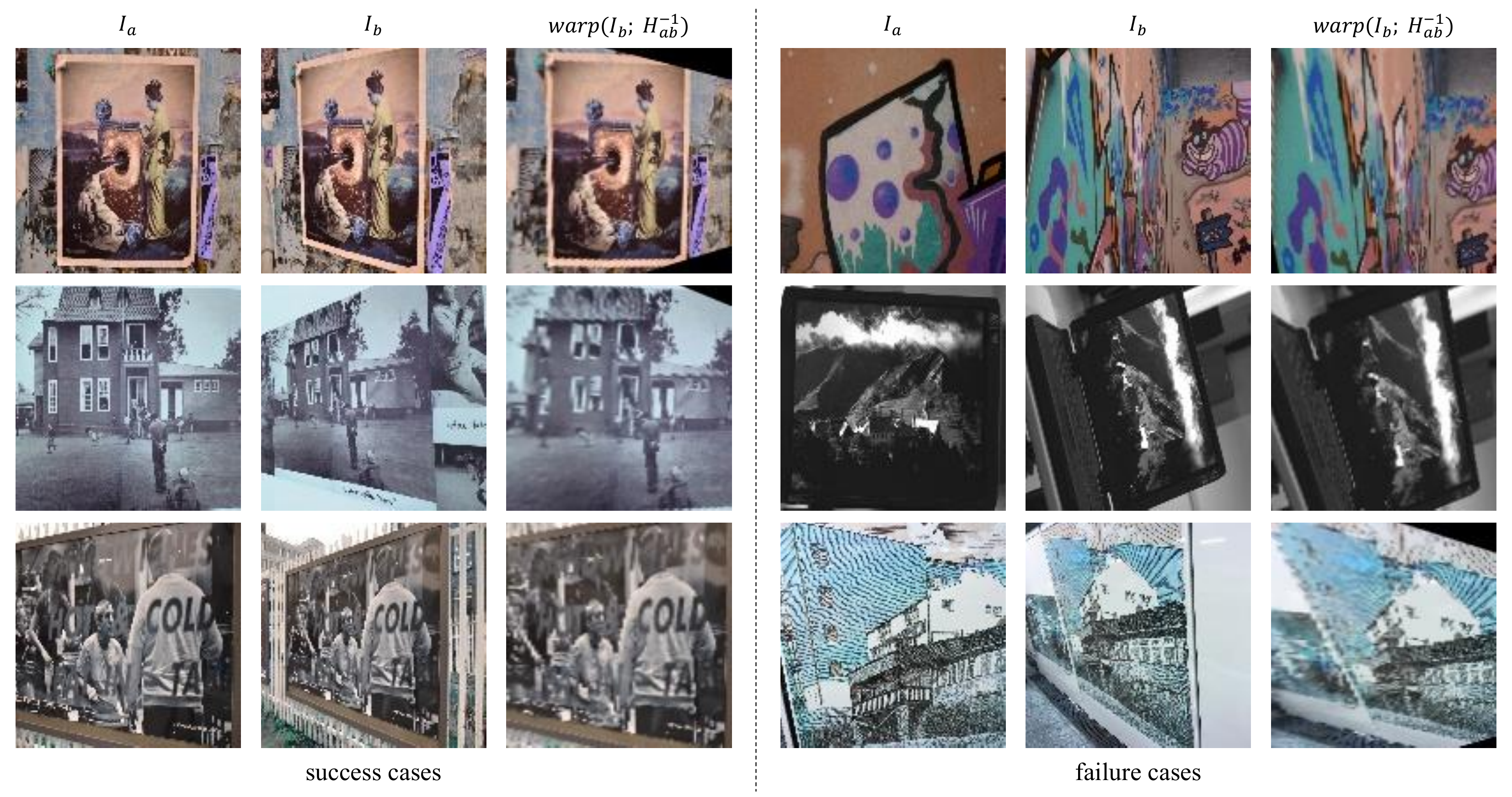}
\caption{Some success and failure cases of our model on the HPatches dataset. We warp the image $I_{b}$ with the inverse of the predicted homography matrix $H_{ab}$. The more accurate the predicted homography matrix, the more aligned $I_{a}$ and the warped image will be.}
\label{fig:cases}
\end{figure*}

As shown in Fig.~\ref{fig:hpatchesresults}, our model achieves a distinct advantages on the HPatches dataset among all deep learning models. However, it is inferior to SIFT and SuperPoint. A major reason may be that the distributions of training images and test images are quite different. The images in the HPatches dataset contains a lot of local illumination variations and non-planar regions. These challenging factors are difficult to simulate by synthetic means. We show some success and failure cases of our model in Fig.~\ref{fig:cases}.

Currently, deep learning models do not achieve satisfactory results on the HPatches dataset. However, they have the advantage of being able to generalize to a variety of scenarios when training on a large number of images, as demonstrated by previous experiments on the synthetic datasets. We believe that the performance of our model, as well as other deep learning models, on the HPatches dataset will be further improved if large-scale real dataset are available for training in the future.

\begin{table}[!t]
\caption{MACE for four variants of our model on the S-COCO dataset. Lower MACE means better performance.}
\centering
\begin{tabular}{|c||c|c|c|c|}
\hline
	& FH & FMH & FMRH-s & FMRH-ss \\
\hline
	MACE & 1.79 & 1.38 & 1.26 & \textbf{0.73} \\
\hline
\end{tabular}
\label{tab:ablation}
\end{table}

\subsection{Ablation Study}
In this section we conduct experiments to investigate the effectiveness of each proposed technique in our model. To this end, we perform ablation experiments on four variants of our model:
\begin{itemize}
	\item \textbf{FH}, which only contains the feature extractor and the homography estimation. It learns the mapping function from image features to the homography matrix.
	\item \textbf{FMH}, which augments \textbf{FH} with a feature matching module. It learns the mapping function from the cost volume produced by the feature matching module to the homography matrix.
	\item \textbf{FMRH-s}, which further augments \textbf{FMH} with an outlier removal module. It mimics the traditional homography estimation pipeline in a single deep learning model. It is trained with the training loss in Eq.~\ref{eq:ls}.
	\item \textbf{FMRH-ss}, has exactly the same network structure as \textbf{FMRH-s}, but is trained with the training loss in Eq.~\ref{eq:lss}.
\end{itemize}

Table~\ref{tab:ablation} reports comparison results on the S-COCO dataset between these four variants. \textbf{FH}, without using the feature matching module and the outlier removal module, is similar to HomographyNet with different feature learning structures. Thus, \textbf{FH} performs similarly to HomographyNet on the S-COCO dataset. The performance comparison between \textbf{FH} and \textbf{FMH} manifests the benefit of leveraging the feature matching module to bridge the gap between image features and the homography matrix. Although \textbf{FMRH-s} is further equiped with an outlier removal module based on \textbf{FMH}, the outlier removal module can not be well trained with the supervised loss in Eq.~\ref{eq:ls}. As a consequence, \textbf{FMRH-s} is only slightly better than \textbf{FMH}. The performance is improved marginally from \textbf{FMRH-s} to \textbf{FMRH-ss}, which is benefited from the proposed self-supervised loss for training the outlier removal module.

Besides the network structure and the loss function, in practice, we find that the backbone of the feature extractor and the outlier removal module also has a significant impact on the performance of our model. Early deep learning models usually employ VGG as the backbone of the feature extractor. However, recent studies~\cite{2020Content,2021Perceptual} have shown that replacing VGG with ResNet-34 always results in better performance in the homography estimation task. Therefore, we choose ResNet-34 as the backbone of the feature extractor in our model. Using the ResNet-34, our model achieves a MACE of 0.73 on the S-COCO dataset. While using the VGG, the MACE is 2.36. For fair comparison, in previous experiments, we implemented all deep learning models to use the same ResNet-34 as the backbone of the feature extractor. As for the outlier removal module,we started with a classic deep denoising model DnCNN~\cite{2017beyond}, but later found that this model performed slightly worse than the currently used model UNet.

\section{Conclusion}
In this work, we have presented a new deep learning model that mimics the traditional homography estimation pipeline. Specifically, the proposed model consists of four components: a feature extractor, a feature matching module, an outlier removal module, and a homography estimator. The feature matching module is implemented using the cost volume technique and has no trainable parameters. The outlier removal module is built based on the UNet structure to remove outliers in the cost volume. We propose a novel self-supervised loss to train the outlier removal module. The entire model can be trained in an end-to-end fashion with a combination of a supervised and the self-supervised loss. Extensive experiments on synthetic and real datasets show that the proposed model significantly outperforms existing deep learning models.

\bibliographystyle{IEEEtran}
\bibliography{SSORN}

\end{document}